# Graphical Representations of Consensus Belief

David M. Pennock and Michael P. Wellman
University of Michigan AI Laboratory
1101 Beal Avenue
Ann Arbor, MI 48109-2110 USA
{dpennock,wellman}@umich.edu

## Abstract

Graphical models based on conditional independence support concise encodings of the subjective belief of a single agent. A natural question is whether the *consensus* belief of a *group* of agents can be represented with equal parsimony. We prove, under relatively mild assumptions, that even if everyone agrees on a common graph topology, no method of combining beliefs can maintain that structure. Even weaker conditions rule out local aggregation within conditional probability tables. On a more positive note, we show that if probabilities are combined with the logarithmic opinion pool (LogOP), then commonly held *Markov* independencies are maintained. This suggests a straightforward procedure for constructing a *consensus Markov network*. We describe an algorithm for computing the LogOP with time complexity comparable to that of exact Bayesian inference.

## 1 Introduction

Suppose that you are charged with the task of modeling the effect of interest rates on the inflation rate. For simplicity, assume that there are only two relevant binary uncertain events: interest rates rise and inflation rates rise. A full joint probability distribution describing this situation would assign a probability to each of the four possible combinations of outcomes. As the number of modeled events increases, the size of the joint distribution grows exponentially. Yet often, the probabilistic relationships can be specified more naturally and compactly in terms of local probabilistic dependencies among events. *Graphical models* offer a language for describing a joint distribution in terms of events and the conditional dependence between them [Jensen, 1996, Pearl, 1988, Whittaker, 1990]. Expert systems based on such models are among the most successful and practical products to emerge from artificial intelligence (AI) research. One of their key features is the ability to efficiently encode an otherwise unmanageably large joint distribution. Indeed, if sufficient conditional independencies (CIs) exist, then memory requirements are exponentially reduced.

From an AI perspective, a graphical model typically encodes the subjective belief of a single agent. We address the more general task of compactly representing the *consensus* or combined belief of a group of agents.

For a modeler, decisions about how to combine beliefs are almost unavoidable. For example, in pursuit of an accurate distribution over interest and inflation rates, you may wish to consult several economists. In a larger model, each expert might be a specialist in some subset of the complete domain. When several related models already exist, it may be desirable to conglomerate their knowledge into a single, more general representation. Even when consulting only one expert to construct only one model, the *designer's* beliefs inevitably play a role— for example, he or she may choose to correct for typical biases of those unfamiliar with probability theory—in fact, choosing *not* to correct for bias itself may distort the expert's true beliefs.

Can the success of graphical models within the single-agent framework be extended to this multiagent setting? More specifically, given each of the agents' beliefs, and some reasonable aggregation rule, will the combined belief have enough structure to warrant a graphical representation?

Decades of research have yielded a variety of prescriptions for aggregating beliefs, which we survey briefly in Section 2. We distinguish between two prevailing methodologies. The first, which enjoys a rich history within statistics and the decision sciences, defines aggregation over joint distributions [Dalkey, 1975, Genest and Zidek, 1986, Wagner, 1984, Madansky, 1964]. The second, more recent and more popular within the AI community, focuses on combining graphical models [Matzkevich and Abramson, 1992, Ng and Abramson, 1994, Xiang, 1996].

This paper demonstrates that common assumptions regard-ing the aggregation of joint distributions imply severe limitations for combining graphical topologies. Most proposals for combining graphical models assume that, at a minimum, if all agents' beliefs conform to a particular structure, then the consensus model should mirror that structure. Yet, as we see in Section 3, almost all proposed statistical aggregation methods violate this property. In fact, we prove that *no* combination function can preserve unanimous structures while simultaneously satisfying other natural and desirable properties. We also demonstrate that essentially no combination function can operate within each conditional probability table separately, and also depend only on the underlying joint distributions. In Section 4, we show that, although it cannot maintain arbitrary structures, a weighted geometric average aggregation rule, called the LogOP, does maintain all unanimously agreed-upon *Markov* structures. We then describe procedures for generating consensus structures that are consistent with the LogOP. Section 5 presents an algorithm for computing the LogOP that, if the consensus structure is sufficiently sparse, can run exponentially faster than a brute force approach.

## 2 Background: Belief and Consensus Belief

Section 2.1 describes background and related work on subjective probability and belief aggregation. In Section 2.2, we cover relevant material on two graphical models for representing probability distributions—Bayesian networks (BNs) and Markov networks (MNs)—and discuss their applicability for encoding both individual and multiagent belief.

### 2.1 Opinion pools and aggregation properties

Suppose that $n$ agents are uncertain about $m$ binary events, $A_1, A_2, \ldots, A_m$, and thus do not know which of the $2^m$ possible joint outcomes or *atomic states* will eventually obtain.[1] Let $Z = \{A_1, A_2, \ldots, A_m\}$ be the set of events, and let $\Omega = \{\omega_1, \omega_2, \ldots, \omega_{2^m}\}$ be the set of all $2^m$ (exclusive, exhaustive) atomic states. We refer to the $A_j$ as the *primary* events, to distinguish them from the other $2^{2^m} - m$ possible sets of atomic states, each of which is also an event. A joint probability distribution Pr associates a probability with each atomic state.

A designer of a probabilistic expert system must assign, implicitly or explicitly, all $2^m$ probabilities. In many situations—for example, when the modeled events encompass a domain broader than any one expert's specialty—more than one source is consulted for probabilities. There is a large body of work in the statistics literature which addresses the aggregation of experts' beliefs into a single, coherent representation.

---

[1] An example of an atomic state is $A_1 \wedge \overline{A}_2 \wedge \overline{A}_3 \wedge \cdots \wedge A_m$ or, written more simply, $A_1 \overline{A}_2 \overline{A}_3 \ldots A_m$.

If each of the $n$ experts, including the system designer, holds a subjective belief $\Pr_i$, then a *consensus* joint probability distribution $\Pr_0$ is any function $f$ of the $\Pr_i$:

$$\Pr_0 \equiv f(\Pr_1, \Pr_2, \ldots, \Pr_n), \qquad (1)$$

where $\Pr_0$ is itself a legal joint probability distribution. The combination function $f$ is often called an *opinion pool*. Many pooling functions over many years have been proposed; Genest and Zidek [1986] provide an excellent overview of the various kinds and discuss their relative merits. The two most common and well-studied are the linear and logarithmic opinion pools (LinOP, LogOP). The LinOP is a weighted arithmetic mean of the members' probabilities for atomic events,

$$\Pr_0(\omega_j) = \sum_{i=1}^{N} \alpha_i \Pr_i(\omega_j), \qquad (2)$$

and the LogOP is a weighted geometric mean,

$$\Pr_0(\omega_j) = \frac{\prod_{i=1}^{N} [\Pr_i(\omega_j)]^{\alpha_i}}{\sum_{k=1}^{2^m} \prod_{i=1}^{N} [\Pr_i(\omega_k)]^{\alpha_i}}, \qquad (3)$$

where the $\alpha_i$, called *expert weights*, are nonnegative numbers that sum to one. A third pooling method identifies one distinguished individual $h$ (real or fictitious, within or outside the group) as a so-called *supra Bayesian* [Lindley, 1985]. The consensus is then defined as the supra Bayesian's posterior distribution, given the "evidence" provided by all of the experts' opinions:

$$\Pr_0(\omega|\Pr_1, \ldots, \Pr_N) \propto \Pr_h(\Pr_1, \ldots, \Pr_N|\omega)\Pr_h(\omega). \qquad (4)$$

Because this approach takes a single agent's perspective, it is well grounded in normative Bayesian theory. Implementing it requires that we choose the supra Bayesian, or assess its prior belief if it is fictitious [Genest and Zidek, 1986]. Computing the posterior further requires that the supra Bayesian specify a joint distribution over all other agents' beliefs.

Attempts to justify more symmetric opinion pools often proceed by posing axioms on the combination function, and arguing that they represent desirable properties [Dalkey, 1975, Genest, 1984c, Genest, 1984b, Genest, 1984a, Genest and Zidek, 1986, Genest et al., 1986, Genest and Wagner, 1987, Wagner, 1984]. Researchers have proved that certain pooling formulae are implied by certain sets of properties. We begin with two seemingly incontrovertible assumptions.

**Property 1 (Unanimity (UNAM))** *If $\Pr_h(\omega) = \Pr_i(\omega)$ for all agents $h$ and $i$, and for all states $\omega \in \Omega$, then $\Pr_0(\omega) = \Pr_1(\omega)$.*

**Property 2 (Nondictatorship (ND))** *There is no single agent i such that* $\Pr_0(\omega) = \Pr_i(\omega)$ *for all* $\omega \in \Omega$, *and regardless of the agents' beliefs.*

UNAM states that if everyone's assessments are in complete agreement, then the consensus agrees as well. ND simply ensures that what is inherently a multiagent problem is not reduced to the single-agent case.

**Property 3 (Marginalization property (MP))** *Let E be an arbitrary event, that is, any subset of* $\Omega$. *Then,*

$$f(\Pr_1, \Pr_2, \ldots, \Pr_n)(E) = f(\Pr_1(E), \Pr_2(E), \ldots, \Pr_n(E)).$$

**Property 4 (Externally Bayesian (EB))** *Let E and F be arbitrary events. Then,*

$$f(\Pr_1, \Pr_2, \ldots, \Pr_n)(E|F) = f(\Pr_1|F, \Pr_2|F, \ldots, \Pr_n|F)(E).$$

MP and EB require consistency for probabilistic operations performed before and after pooling. MP states that we obtain the same probability for an event $E$ whether we pool the opinions first, and then compute $\Pr_0(E) = \sum_{\omega \in E} \Pr_0(\omega)$, or if we first compute $\Pr_i(E) = \sum_{\omega \in E} \Pr_i(\omega)$ for each agent $i$, and then pool their opinions only over $E$. Similarly, EB holds that we obtain the same $\Pr_0(E|F)$ whether we combine opinions first and condition on $F$ second, or condition on $F$ first and combine opinions second. It has been shown that any $f$ satisfying both MP and UNAM is a LinOP [Genest, 1984c], and any satisfying EB and UNAM is a LogOP [Genest, 1984a]. Genest [1984b] also shows that $f$ cannot simultaneously satisfy MP, EB, UNAM, and ND.

**Property 5 (Proportional dependence on states (PDS))**

$$\Pr_0(\omega) \propto f(\Pr_1(\omega), \Pr_2(\omega), \ldots, \Pr_n(\omega)).$$

PDS is sometimes called *independence of irrelevant states*, or termed a *likelihood principle*. It assures that the consensus likelihood ratio between two states does not depend on the agents' assessments of any other "irrelevant" state. The LinOP, LogOP, and most other proposed opinion pools satisfy PDS.

**Property 6 (Independence preservation property (IPP))** *Let E and F be arbitrary events. If* $\Pr_i(E|F) = \Pr_i(E)$ *for all agents i, then* $\Pr_0(E|F) = \Pr_0(E)$.

IPP requires that *all* unanimously held independencies are preserved in the consensus. Advocates of IPP reason that identifying the independencies in a model is central to understanding the underlying phenomena, and that complete agreement on this dimension should be embraced. On the other hand, Genest and Wagner [1987] make a compelling case *against* the use of IPP by proving that *no aggregation function whatsoever can satisfy it* along with PDS and ND, when $|\Omega| \geq 5$.

One might argue that IPP is overly strong. It requires preservation of, for example, a unanimous independence between the events $E = A_3 \overline{A}_7$ and $F = \overline{A}_2 A_4 \vee A_7$. This kind of independence seems of little descriptive value to a modeler, and indeed cannot be represented with a BN. The designer may be willing to forgo preserving *all* independencies, being content to preserve independencies among the *primary* events, $A_1, A_2, \ldots, A_m$. With this in mind, we define a weaker independence property.

**Property 7 (Event independence preservation property (EIPP))** *If* $\Pr_i(A_j|A_k) = \Pr_i(A_j)$ *for all agents i, then* $\Pr_0(A_j|A_k) = \Pr_0(A_j)$.

In Section 3, we see that substituting EIPP for IPP does admits a possibility that is consistent with both PDS and ND, though not a very satisfactory one. In search of a nontrivial possibility, we define two even weaker independence conditions.

**Property 8 (Markov event independence preservation property (MEIPP))** *If* $\Pr_i(A_j|WA_k) = \Pr_i(A_j|W)$ *for all agents i and for all* $W \subseteq Z$ *(including* $W = \emptyset$*), then* $\Pr_0(A_j|A_k) = \Pr_0(A_j)$.

**Property 9 (Non-Markov event independence preservation property (NMEIPP))** *If* $\Pr_i(A_j|A_k) = \Pr_i(A_j)$ *for all agents i, and* $\Pr_h(A_j|WA_k) \neq \Pr_h(A_j|W)$, *for some agent h and some* $W \subseteq Z$, *then* $\Pr_0(A_j|A_k) = \Pr_0(A_j)$.

These two properties are purposely constructed so that EIPP $\Leftrightarrow$ (MEIPP $\wedge$ NMEIPP). We see in Section 3 that the source of the impossibility lies entirely within the latter. Finally, we define a stronger version of the MEIPP.

**Property 10 (Markov independence preservation property (MIPP))** *Let* $W, X \subseteq Z - A_j$ *be disjoint sets of events such that* $A_j \cup W \cup X = Z$. *If* $\Pr_i(A_j|WX) = \Pr_i(A_j|W)$ *for all agents i, then* $\Pr_0(A_j|WX) = \Pr_0(A_j|W)$.

The relative strengths of these various independence conditions can be summarized as follows:

$$\text{IPP} \Rightarrow \text{EIPP} \Leftrightarrow (\text{MEIPP} \wedge \text{NMEIPP})$$
$$\text{MIPP} \Rightarrow \text{MEIPP}$$

### 2.2 Graphical models for belief and consensus belief

The *Bayesian network* (BN) has proved invaluable as a language for compactly encoding a joint probability distribution [Jensen, 1996]. Conciseness is achieved by factoring atomic states into primary events, and exploiting conditional independence among these events. Consider the event $A_k \in Z$, with events $A_1, A_2, \ldots, A_{k-1}$ preceding it in index order. Suppose that, given the outcomes of a subset $\mathbf{pa}(A_k) \subseteq \{A_1, A_2, \ldots, A_{k-1}\}$ of its preceding events—called $A_k$'s *parents*—the event $A_k$ is conditionally independent of all other preceding events. This structure can be depicted graphically as a *directed acyclic graph*: each event is a node in the graph, and there is a directed edge from node $A_j$ to node $A_k$ if and only if $A_j$ is a parent of $A_k$. We also refer to $A_k$ as the *child* of $A_j$, and $A_k \cup \mathbf{pa}(A_k)$ as the *family* of $A_k$. We can write the joint probability distribution in a (usually) more compact form:

$$\Pr(A_1 A_2 \cdots A_m) = \prod_{k=1}^{m} \Pr(A_k | \mathbf{pa}(A_k)).$$

For each event $A_k$, we record a *conditional probability table* (CPT), which contains probabilities $\Pr(A_k|\mathbf{pa}(A_k))$ for all possible combinations of outcomes of events in $\mathbf{pa}(A_k)$. Thus it is possible to implicitly represent the full joint with $O(2^q m)$ probabilities, instead of $O(2^m)$, where $q$ is the maximum number of parents of any node in the network.

A *Markov network* (MN) is another graphical language for modeling conditional independence and for implicitly describing a joint distribution [Whittaker, 1990, Darroch et al., 1980]. Events are again associated with nodes in a graph, and edges encode probabilistic dependencies. However, as opposed to BNs, the underlying structure of a MN is an *undirected* graph. Given the outcomes of its direct neighbors, an event $A_j$ is conditionally independent of *every* other event in the network, not just preceding events. The neighbors of an event form a *Markov blanket* around it, "shielding" it from direct influence from the rest of the events [Pearl, 1988]. We call the node $A_j$ and the set of nodes $X \subseteq Z - A_j$ *Markov independent*, given another set $W \subseteq Z - X - A_j$, if $\Pr(A_j|WX) = \Pr(A_j|W)$ and $A_j \cup W \cup X = Z$. Thus a node is Markov independent of all other nodes, given its blanket. Encoding the joint distribution implied by a MN involves assigning a *potential* probability to each clique [Neapolitan, 1990, Pearl, 1988].

The Markov blanket of a node in a BN consists of its direct parents, its direct children, and its children's direct parents [Pearl, 1988]. Therefore a BN can be converted into a MN by *moralizing* the network, or fully connecting ("marrying") each node's parents, and dropping edge directionality [Lauritzen and Spiegelhalter, 1988, Neapolitan, 1990]. A MN can be converted into a BN by *filling in* or *triangulating* [Kloks, 1994] the graph, and adding directionality according to the fill-in ordering [Jensen, 1996, Lauritzen and Spiegelhalter, 1988, Neapolitan, 1990, Pearl, 1988]. Both transformations are sound with respect to independence, but neither is complete. A filled-in BN is also called *decomposable* [Chyu, 1991, Darroch et al., 1980, Pearl, 1988, Shachter et al., 1991].

Although most BN research concerns modeling a single agent's belief, some researchers have examined the use of BNs in a multiagent context. Ng and Abramson [1994] describe an architecture called the *probabilistic multi-knowledge-base system*, which consists of a collection of BNs, each encoding the knowledge of a single expert. The authors choose to keep the BNs separate and combine probabilities *at run time* with a variable-weight variant of the LinOP. They address a variety of engineering issues, including the elicitation and propagation of expert confidence information, and build a working prototype to diagnose pathologies of the lymph system. Xiang [1996] describes conditions under which *multiply sectioned Bayesian networks*, originally developed for single agent reasoning, can represent the combined beliefs of multiple agents. The main assumption is that, whenever two agents' BNs contain some of the same events, they must agree on the joint distribution over these common events. Bonduelle [1987] prescribes both normative and behavioral techniques for a decision maker (DM) to identify and reconcile differences of opinion among experts. When those opinions are expressed as graphical models, he suggests that the DM first choose a consensus topology, and then calculate aggregate probabilities. Jacobs [1995] compares the LinOP and supra Bayesian approaches as methods for combining the multiple feature analyzers found in real and artificial neural systems.

Matzkevich and Abramson [1992] give an algorithm for explicitly combining two BN DAGs into a single DAG, or *fusing* the two topologies. The algorithm transfers one arc at a time from the second DAG to the first, possibly reversing the arc in order to remain consistent with the current partial ordering. Reversing arcs may add new arcs to the second DAG [Shachter, 1988], which would in turn need to be transferred. In a second paper, the same authors show [1993] that the task of minimizing the number of arcs in their combined DAG is NP-hard, as are several other related tasks. They argue that, intuitively, the consensus model should capture independencies agreed upon by at least $c \leq n$ of the agents; in particular, when $c = n$ and the orderings are mutually consistent, the consensus DAG should be a union of the individual DAGs. In both of these papers, and in Bonduelle's work, it is essentially assumed that the EIPP, or a stronger version thereof, should hold.

Though Matzkevich and Abramson make no commitment on how to combine probabilities, they do give an example [1992] where the LinOP is applied *locally*, or separately within each CPT. We say that such a localized aggregator satisfies the *family aggregation* (FA) property.

**Property 11 (Family aggregation (FA))**

$\Pr_0(A_j|\mathbf{pa}(A_j)) = $
    $f(\Pr_1(A_j|\mathbf{pa}(A_j)), \ldots, \Pr_n(A_j|\mathbf{pa}(A_j))).$

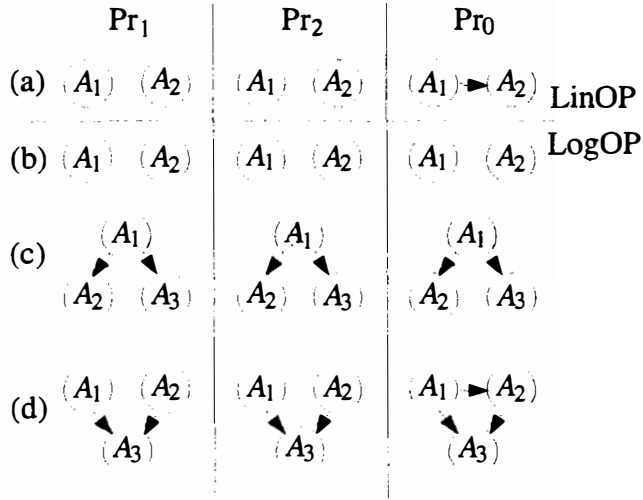

Figure 1: Independence preservation behavior of (a) LinOP and (b)–(d) LogOP. If two agents' beliefs $\text{Pr}_1$ and $\text{Pr}_2$ have the dependency structures shown, then the consensus $\text{Pr}_0$ will in general have the dependency structure depicted in column three.

Although FA may seem natural, we see in Section 3 that it conflicts with other compelling properties.

## 3    Combining Bayesian Networks: Examples and Impossibility

In Sections 3.1 and 3.2 we consider the implications of the properties EIPP and FA, respectively.

### 3.1    Event Independence Preservation

**Example 1 (EIPP and the LinOP)**

Suppose that two agents agree that two primary events, $A_1$ and $A_2$, are independent, as pictured in Figure 1(a), but disagree on the associated marginal probabilities. For concreteness, let the first agent hold beliefs $\text{Pr}_1(A_1) = \text{Pr}_1(A_2) = 0.5$, and the second $\text{Pr}_2(A_1) = 0.8$ and $\text{Pr}_2(A_2) = 0.6$. Thus,

$$\text{Pr}_1(A_1 A_2) = 0.25 \qquad \text{Pr}_2(A_1 A_2) = 0.48$$
$$\text{Pr}_1(A_1 \overline{A}_2) = 0.25 \qquad \text{Pr}_2(A_1 \overline{A}_2) = 0.32$$
$$\text{Pr}_1(\overline{A}_1 A_2) = 0.25 \qquad \text{Pr}_2(\overline{A}_1 A_2) = 0.12$$
$$\text{Pr}_1(\overline{A}_1 \overline{A}_2) = 0.25 \qquad \text{Pr}_2(\overline{A}_1 \overline{A}_2) = 0.08.$$

Now if we apply the LinOP (2) with, say, equal weights of $w_1 = w_2 = 0.5$, we get:

$$\text{Pr}_0(A_1 A_2) = 0.365$$
$$\text{Pr}_0(A_1 \overline{A}_2) = 0.41$$
$$\text{Pr}_0(\overline{A}_1 A_2) = 0.185$$
$$\text{Pr}_0(\overline{A}_1 \overline{A}_2) = 0.165.$$

In particular, $\text{Pr}_0(A_1)\text{Pr}_0(A_2) \neq \text{Pr}_0(A_1 A_2)$, and so the two events are *not* independent in the consensus.[2] Even though the precondition of the EIPP is met, the postcondition is not: a BN representation of the derived consensus would have to include an edge between $A_1$ and $A_2$. □

**Example 2 (EIPP and the LogOP)**

Suppose that two agents' beliefs over two primary events are as described in Example 1. If we apply the LogOP with equal weights, we get:

$$\text{Pr}_0(A_1 A_2) = 0.367007$$
$$\text{Pr}_0(A_1 \overline{A}_2) = 0.29966$$
$$\text{Pr}_0(\overline{A}_1 A_2) = 0.183503$$
$$\text{Pr}_0(\overline{A}_1 \overline{A}_2) = 0.14983.$$

In this case, $\text{Pr}_0(A_1)\text{Pr}_0(A_2) = \text{Pr}_0(A_1 A_2)$, and the two events remain independent, as shown in Figure 1(b). This is not a numerical coincidence; in fact, independence between only two events is always maintained by the LogOP [Genest and Wagner, 1987]. Now suppose that among three primary events, both agents agree that $A_3$ is independent of $A_2$ given $A_1$. That is, both agents agree that dependencies conform to a tree structure, with $A_1$ the parent of both $A_2$ and $A_3$, as depicted in Figure 1(c). Then once again, the LogOP will maintain this structure. One might conjecture that the LogOP maintains all BN structures, but this is not the case. For example, suppose that, among three primary events, the two agents agree that $A_1$ and $A_2$ are mutually independent, and that $A_3$ depends on both $A_1$ and $A_2$. That is, both agents agree on the polytree structure in Figure 1(d). In this case, when we compute the consensus with the LogOP, $A_1$ and $A_2$ will in general become mutually dependent, the EIPP is not satisfied, and a consensus BN will require an arc between the two nodes. □

Having seen that both the LinOP and the LogOP violate the EIPP, we seek a more general characterization of the class of functions that do obey it. We begin by showing that Lemma 3.2 in [Genest and Wagner, 1987], originally proved with respect to the IPP, is also applicable under the weaker EIPP.

**Lemma 1 (Adapted from [Genest and Wagner, 1987])**
*If $f$ obeys EIPP and PDS, then there exist constants $\alpha_1, \alpha_2, \ldots, \alpha_n$, and $c$ such that*

$$\text{Pr}_0(\omega_j) = \sum_{i=1}^n \alpha_i \text{Pr}_i(\omega_j) + c. \qquad (5)$$

---
[2] As early as Yule [1903] it was recognized that averaging two distributions may mask a commonly held independence.

**Proof (sketch).** Consider three events $A_1$, $A_2$, and $A_3$, with agents' beliefs described as follows:

$$\Pr_i(A_1 A_2 A_3) = \Pr_i(A_1 A_2 \overline{A}_3) = \frac{(1-z_i)^2}{4(1+z_i)}$$

$$\Pr_i(A_1 \overline{A}_2 A_3) = \Pr_i(A_1 \overline{A}_2 \overline{A}_3) = \frac{1-z_i}{4}$$

$$\Pr_i(\overline{A}_1 \overline{A}_2 A_3) = x_i$$

$$\Pr_i(\overline{A}_1 \overline{A}_2 \overline{A}_3) = y_i, \quad (6)$$

where $z_i = x_i + y_i$ for all $i$. In this case, all agents agree that $A_1$ and $A_2$ are independent and, as long as $z_i < 1$, these equations describe a legal probability distribution. Since $f$ obeys PDS, there must be some function $g$ such that,

$$\Pr_0(\overline{A}_1 \overline{A}_2 A_3) = \frac{g(x_1, x_2, \ldots, x_n)}{\sum_{k=1}^{8} g(\Pr_1(\omega_k), \ldots, \Pr_n(\omega_k))}$$

and similarly for $\Pr_0(\overline{A}_1 \overline{A}_2 \overline{A}_3)$. Now imagine a second situation exactly as in (6), except with $\Pr_i(\overline{A}_1 \overline{A}_2 A_3) = x_i'$ and $\Pr_i(\overline{A}_1 \overline{A}_2 \overline{A}_3) = y_i'$. Genest and Wagner show that, as long as $x_i + y_i = x_i' + y_i' < 1$, then

$$g(x_1, x_2, \ldots, x_n) + g(y_1, y_2, \ldots, y_n)$$
$$= g(x_1', x_2', \ldots, x_n') + g(y_1', y_2', \ldots, y_n'). \quad (7)$$

From here, they show that *since $x_i$ and $y_i$ can be chosen arbitrarily* (as long as their sum is less than one), then $f$ must have the form specified. □

Genest and Wagner go on to show, without further assumption, that $f$ must be a dictatorship. However, that proof does *not* carry through under the weaker condition EIPP. This can be seen via a simple counterexample. Let $f$ always ignore the agents' opinions, and simply assign a uniform distribution over all $\omega \in \Omega$. In this case, the consensus distribution holds that *all* primary events $A_j$ are independent, and thus any agreed upon independencies are trivially maintained. One might wonder whether EIPP admits any other, more appealing, aggregation functions. The following proposition essentially establishes that it does not.

**Proposition 1** *No aggregation function $f$ can simultaneously satisfy EIPP, PDS, UNAM, and ND.*

**Proof.** With the addition of UNAM, it is clear that $c$ must be zero in (5), and thus $f$ must have the form of a standard LinOP (2). From Example 1, we know that the LinOP does not maintain independence even between just two events. The fact that the LinOP cannot satisfy both IPP and ND is proved formally by several authors [Genest, 1984c, Lehrer and Wagner, 1983, Wagner, 1984]. Their proofs are applicable to EIPP as well, since they hold even when $|\Omega| = 4$, in which case EIPP and IPP coincide. □

A careful examination of the proof of Lemma 1 also suggests one more possibility when the full generality of IPP is relaxed. Suppose that all agents agree that *all three* events, $A_1$, $A_2$, and $A_3$, are completely independent. Then it can be shown that $\Pr_i(\overline{A}_1 A_2 A_3) = z_i/(1+z_i) + y_i$ and, furthermore, that $x_i = y_i$ for all $i$. In this case, (7) holds only vacuously, since $x_i' = x_i$ and $y_i' = y_i$. Moreover, since $x_i$ and $y_i$ are no longer arbitrary, the proof does not go through. Thus, under this fully independent condition, the conclusion of Lemma 1 is no longer valid.

This insight leads us to characterize the inherent impossibility more sharply, by dividing EIPP into two, weaker conditions, NMEIPP and MEIPP, and showing that the former retains the impossibility while the latter does not.

**Corollary 1** *No aggregation function $f$ can simultaneously satisfy NMEIPP, PDS, UNAM, and ND.*

**Proof.** The proof of Lemma 1 still follows under NMEIPP, and thus so does the proof of Proposition 1. □

Section 4 demonstrates that in fact, MEIPP is perfectly consistent with PDS, UNAM, and ND in a nontrivial way. Indeed, the stronger MIPP is consistent as well.

### 3.2 Family Aggregation

**Example 3 (Family aggregation)**

Consider two agents, each with a BN consisting of two primary events, with $A_1$ the parent of $A_2$ and with beliefs as follows:

$$\Pr_1(A_1) = 0.2 \qquad \Pr_2(A_1) = 0.8$$
$$\Pr_1(A_2|A_1) = 0.4 \qquad \Pr_2(A_2|A_1) = 0.8$$
$$\Pr_1(A_2|\overline{A}_1) = 0.6 \qquad \Pr_2(A_2|\overline{A}_1) = 0.3$$

We compute each consensus CPT as an average of the corresponding individual CPTs. That is, $\Pr_0(A_1) = (.2 + .8)/2 = .5$, $\Pr_0(A_2|A_1) = (.4 + .8)/2 = .6$, etc. This results in the following consensus joint distribution:

$$\Pr_0(A_1 A_2) = 0.3$$
$$\Pr_0(A_1 \overline{A}_2) = 0.2$$
$$\Pr_0(\overline{A}_1 A_2) = 0.225$$
$$\Pr_0(\overline{A}_1 \overline{A}_2) = 0.275.$$

Next suppose that both agents reverse their edge between the two events, such that $A_2$ is the parent of $A_1$, but that their joint distributions remain unchanged. Now the agents' CPTs are:

$$\Pr_1(A_2) = 0.56 \qquad \Pr_2(A_2) = 0.7$$
$$\Pr_1(A_1|A_2) = 0.142857 \qquad \Pr_2(A_1|A_2) = 0.914286$$
$$\Pr_1(A_1|\overline{A}_2) = 0.272727 \qquad \Pr_2(A_1|\overline{A}_2) = 0.533333$$

and if we average locally within each CPT, we get a different consensus distribution:

$$\text{Pr}_0(A_1 A_2) = 0.333$$
$$\text{Pr}_0(A_1 \overline{A}_2) = 0.149121$$
$$\text{Pr}_0(\overline{A}_1 A_2) = 0.297$$
$$\text{Pr}_0(\overline{A}_1 \overline{A}_2) = 0.220878.$$

Thus averaging only within each family of the BN violates the form of the opinion pool itself (1), which insists that the consensus joint distribution depend only on the underlying joint distributions of the agents involved. □

We now show that this inconsistency is not confined solely to the averaging aggregator.

**Proposition 2** *No aggregation function $f$ can simultaneously satisfy FA, UNAM, and ND.*

**Proof (sketch).** Let the first event in the consensus BN be $A_{j_1}$, the second $A_{j_2}$, ..., and the last $A_{j_m}$. The FA property requires both of the following:

$$\text{Pr}_0(A_{j_1})$$
$$= f(\text{Pr}_1(A_{j_1}), \text{Pr}_2(A_{j_1}), \ldots, \text{Pr}_n(A_{j_1})) \quad (8)$$
$$\text{Pr}_0(A_{j_m}|Z - A_{j_m})$$
$$= f(\text{Pr}_1(A_{j_m}|Z - A_{j_m}), \ldots, \text{Pr}_n(A_{j_m}|Z - A_{j_m})) \quad (9)$$

By the definition of an opinion pool (1), the consensus belief depends only on the agents' underlying joint distributions, and not on the particular ordering of events in each BN. Thus, we must arrive at the same consensus distribution as long as $\{j_1, j_2, \ldots, j_m\}$ is some permutation of $\{1, 2, \ldots, m\}$. Consider two permutations, one where $j_1 = 1$ and one where $j_m = 1$. Then (8) and (9) become:

$$\text{Pr}_0(A_1)$$
$$= f(\text{Pr}_1(A_1), \text{Pr}_2(A_1), \ldots, \text{Pr}_n(A_1)) \quad (10)$$
$$\text{Pr}_0(A_1|Z - A_1)$$
$$= f(\text{Pr}_1(A_1|Z - A_1), \ldots, \text{Pr}_n(A_1|Z - A_1)) \quad (11)$$

Dalkey [1975] proves that no function can simultaneously satisfy (10), (11), UNAM, and ND. Alternatively, the two equations essentially require that $f$ satisfy both MP and EB, which Genest [1984b] shows are incompatible with UNAM and ND. □

## 4  The LogOP and Consensus Markov Networks

The results in Section 3 suggest that insisting upon general event independence preservation has rather severe consequences. In this section, we see that preserving *Markov* independencies is in fact compatible with PDS, UNAM, and ND. Let $A_j$ be a primary event, and $W \subseteq Z - A_j$ and $X = Z - W - A_j$ be sets of events. Then $A_j$ is Markov independent of $X$ given $W$ if $\text{Pr}(A_j|WX) = \text{Pr}(A_j|W)$.

**Proposition 3** *The LogOP satisfies MIPP.*

**Proof.** Since the LogOP is defined in terms of atomic states $\omega$, we make use of the following two identities:

$$\text{Pr}_0(A|WX) \equiv \frac{\text{Pr}_0(AWX)}{\text{Pr}_0(AWX) + \text{Pr}_0(\overline{A}WX)}$$
$$\text{Pr}_0(A|W) \equiv \frac{\sum_X \text{Pr}_0(AWX)}{\sum_X \text{Pr}_0(AWX) + \sum_X \text{Pr}_0(\overline{A}WX)}$$

where $\sum_X$ represents a sum over all possible combinations of outcomes of events in the set $X$. Then we have that,

$$\text{Pr}_0(A|WX) = \frac{\prod_{i=1}^N [\text{Pr}_i(AWX)]^{\alpha_i}}{\prod_{i=1}^N [\text{Pr}_i(AWX)]^{\alpha_i} + \prod_{i=1}^N [\text{Pr}_i(\overline{A}WX)]^{\alpha_i}}$$
$$= \frac{\prod \left[\frac{\text{Pr}_i(AW)\text{Pr}_i(WX)}{\text{Pr}_i(W)}\right]^{\alpha_i}}{\prod \left[\frac{\text{Pr}_i(AW)\text{Pr}_i(WX)}{\text{Pr}_i(W)}\right]^{\alpha_i} + \prod \left[\frac{\text{Pr}_i(\overline{A}W)\text{Pr}_i(WX)}{\text{Pr}_i(W)}\right]^{\alpha_i}}$$
$$= \frac{\prod [\text{Pr}_i(AW)]^{\alpha_i}}{\prod [\text{Pr}_i(AW)]^{\alpha_i} + \prod [\text{Pr}_i(\overline{A}W)]^{\alpha_i}}$$
$$= \frac{\prod [\text{Pr}_i(AW)]^{\alpha_i}}{\prod [\text{Pr}_i(AW)]^{\alpha_i} + \prod [\text{Pr}_i(\overline{A}W)]^{\alpha_i}} \cdot \frac{\sum_X \prod [\text{Pr}_i(WX)]^{\alpha_i}}{\sum_X \prod [\text{Pr}_i(WX)]^{\alpha_i}}$$
$$= \frac{\sum_X \prod [\text{Pr}_i(AW)\text{Pr}_i(WX)]^{\alpha_i}}{\sum_X \prod [\text{Pr}_i(AW)\text{Pr}_i(WX)]^{\alpha_i} + \sum_X \prod [\text{Pr}_i(\overline{A}W)\text{Pr}_i(WX)]^{\alpha_i}}$$
$$= \frac{\sum_X \prod \left[\frac{\text{Pr}_i(AW)\text{Pr}_i(WX)}{\text{Pr}_i(W)}\right]^{\alpha_i}}{\sum_X \prod \left[\frac{\text{Pr}_i(AW)\text{Pr}_i(WX)}{\text{Pr}_i(W)}\right]^{\alpha_i} + \sum_X \prod \left[\frac{\text{Pr}_i(\overline{A}W)\text{Pr}_i(WX)}{\text{Pr}_i(W)}\right]^{\alpha_i}}$$
$$= \frac{\sum_X \prod [\text{Pr}_i(AWX)]^{\alpha_i}}{\sum_X \prod [\text{Pr}_i(AWX)]^{\alpha_i} + \sum_X \prod [\text{Pr}_i(\overline{A}WX)]^{\alpha_i}}$$
$$= \frac{\sum_X \text{Pr}_\bullet(AWX)}{\sum_X \text{Pr}_0(AWX) + \sum_X \text{Pr}_0(\overline{A}WX)}$$
$$= \text{Pr}_0(A|W)$$
□

Suppose that each agent's belief is given as a MN, and we wish to generate a consensus MN structure that can encode the results of the LogOP. As discussed in Section 2.2, graph connectivity in a MN represents probabilistic dependence, and the neighborhood relation represents direct influence. For each node $A_j$, the set of its neighbors plays the role of $W$ in Proposition 3, and all other nodes constitute the set $X$. The proposition ensures that, if all agents agree on a common MN structure, then the consensus distribution derived by the LogOP will respect the same structure. When agents are not in complete agreement on the structure, then the consensus can be represented as a MN defined by the union of all the individual MNs. In other words, there is an edge between $A_j$ and $A_k$ in the consensus MN if and only if there is an edge between those two nodes in at least one of the agents' MNs.

Pearl [1988] gives axiomatic descriptions of both MNs and BNs. Only the former includes an axiom called *strong union*, which states that if $\text{Pr}(A_j|A_k) = \text{Pr}(A_j)$, then $\text{Pr}(A_j|WA_k) = \text{Pr}(A_j|W)$ for all $W \subseteq Z$. Notice that, if the precondition of the EIPP is met, and strong union holds for all agents, then the precondition of the MEIPP must also hold. This axiom is the key distinction that allows common MN structures to be maintained in the LogOP consensus, whereas common BN structures in general are not.

Given a collection of BNs, generating a consensus BN structure that is consistent with the LogOP is also relatively straightforward. We first convert each BN into a MN by moralizing the graphs, or fully connecting each node's parents and dropping edge directionality [Lauritzen and Spiegelhalter, 1988, Neapolitan, 1990]. Next, we compute the union of the individual MNs, and finally we convert the resulting consensus MN back into a BN by filling in or triangulating the network, reintroducing directionality according to the fill-in order[3] [Jensen, 1996, Lauritzen and Spiegelhalter, 1988, Neapolitan, 1990, Pearl, 1988].

We have outlined how to derive consensus MN or BN *structures*; what of computing the associated probabilities? In Section 5, we give an algorithm for computing the probabilities in a consensus BN that is polynomial in the size of its CPTs. Note that, even when all agents agree on a BN structure, the size of the final representation may grow exponentially during fill-in, and computing the union of the intermediate MNs when agents disagree will only exacerbate this problem. Nevertheless, even a decomposable representation can be exponentially smaller than the full joint distribution, and the most popular algorithms for exact Bayesian inference do operate on decomposable models in practice.

## 5  Computing LogOP and LinOP

Since the LinOP (2) and LogOP (3) are defined over atomic states, computing, for example, the consensus marginal probability of a single event involves in the worst case a summation over $2^{m-1}$ terms. Moreover, even computing the LogOP consensus for a single state requires a normalization factor that is itself a sum over all $2^m$ states. In this section, we see that if each agent's belief is represented as a BN, the LinOP and LogOP consensus for any probabilistic query can be computed more efficiently. In particular, for the LogOP, we can compute the CPTs of a consensus BN with time complexity $O(nm^2 2^q)$, where $q$ is the maximum number of parents of any node in the consensus structure.

We focus first on the task of generating a LogOP-consistent consensus BN. We compute its structure as described in Section 4. Consider computing the CPT at $A_j$, that is, $\Pr_0(A_j|\mathbf{pa}(A_j))$ for all combinations of outcomes of events in $\mathbf{pa}(A_j)$. From Proposition 2, we know that simply combining each agent's assessment of this conditional probability will not succeed in general. However, we *can* compute the *last* CPT, $\Pr_0(A_m|\mathbf{pa}(A_m))$, in terms of only the $\Pr_i(A_m|\mathbf{pa}(A_m))$, by computing the LogOP over the single event $A_m$:

$$\Pr_0(A_m|\mathbf{pa}(A_m)) = $$
$$\frac{\prod_{i=1}^{N}[\Pr_i(A_m|\mathbf{pa}(A_m))]^{\alpha_i}}{\prod[\Pr_i(A_m|\mathbf{pa}(A_m))]^{\alpha_i} + \prod[\Pr_i(\overline{A_m}|\mathbf{pa}(A_m))]^{\alpha_i}}. \quad (12)$$

[3]We do not claim that these consensus structures are correct or minimal in any sense, or even that LogOP is the preferred aggregation method. Our goal is more to guide a modeler's decision process by delineating what representations are consistent under what circumstances.

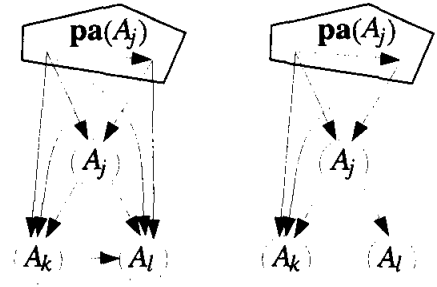

Figure 2: Two potential sections of a decomposable BN. $A_j$'s children can be either in the same clique or in separate cliques.

Because the LogOP satisfies EB, if we condition on *all* other events $Z - A_m$ in the network, then the LogOP over just $A_m$ will return the same result as if we had computed the LogOP over all events, and then conditioned on $Z - A_m$. Equation 12 also reflects the fact that $\Pr_0(A_m|\mathbf{pa}(A_m)) = \Pr_0(A_m|Z - A_m)$ and $\Pr_i(A_m|\mathbf{pa}(A_m)) = \Pr_i(A_m|Z - A_m)$, by the semantics of the BNs.

We can compute the remainder of the CPTs in reverse index order. Assume that the CPTs $\Pr_0(A_k|\mathbf{pa}(A_k))$ have been calculated for all $k > j$, and that next we need to calculate $\Pr_0(A_j|\mathbf{pa}(A_j))$. To simplify the discussion, let $A_j$ have exactly two children, $A_k$ and $A_l$, with $j < k < l$; the analysis generalizes easily to more children (or one child). Since the BN is decomposable, its topology is a tree of cliques [Chyu, 1991, Pearl, 1988, Shachter et al., 1991], and $A_k$ and $A_l$ can either be in the same clique or in separate cliques, as depicted in Figure 2. Note that decomposability also ensures that $A_j$'s neighbors, $A_l \cup A_k \cup \mathbf{pa}(A_j)$, constitute its Markov blanket. We can query each of the agent's BNs for the probabilities $\Pr_i(A_j|A_l \cup A_k \cup \mathbf{pa}(A_j))$ using a standard BN inference algorithm. From these, we can compute the corresponding consensus probability as a LogOP only over $A_j$, as before:

$$\Pr_0(A_j|A_l \cup A_k \cup \mathbf{pa}(A_j))$$
$$\propto \prod_{i=1}^{N}[\Pr_i(A_j|A_l \cup A_k \cup \mathbf{pa}(A_j))]^{\alpha_i}.$$

We now need only eliminate the conditioning on $A_l$ and

$A_k$. By Bayes's rule, we have that

$$\frac{\text{Pr}_0(A_j|A_l \cup A_k \cup \mathbf{pa}(A_j))}{\text{Pr}_0(\overline{A_j}|A_l \cup A_k \cup \mathbf{pa}(A_j))}$$

$$= \frac{\text{Pr}_0(A_l \cup A_k|A_j \cup \mathbf{pa}(A_j))}{\text{Pr}_0(A_l \cup A_k|\overline{A_j} \cup \mathbf{pa}(A_j))} \cdot \frac{\text{Pr}_0(A_j|\mathbf{pa}(A_j))}{\text{Pr}_0(\overline{A_j}|\mathbf{pa}(A_j))}$$

$$= \frac{\text{Pr}_0(A_l|A_k \cup A_j \cup \mathbf{pa}(A_j))}{\text{Pr}_0(A_l|A_k \cup \overline{A_j} \cup \mathbf{pa}(A_j))} \cdot \frac{\text{Pr}_0(A_k|A_j \cup \mathbf{pa}(A_j))}{\text{Pr}_0(A_k|\overline{A_j} \cup \mathbf{pa}(A_j))}$$

$$\cdot \frac{\text{Pr}_0(A_j|\mathbf{pa}(A_j))}{\text{Pr}_0(\overline{A_j}|\mathbf{pa}(A_j))}.$$

Because the BN is decomposable, and regardless of whether $A_k$ and $A_l$ are in the same or different cliques, $\text{Pr}_0(A_l|A_k \cup A_j \cup \mathbf{pa}(A_j)) = \text{Pr}_0(A_l|\mathbf{pa}(A_l))$ and $\text{Pr}_0(A_k|A_j \cup \mathbf{pa}(A_j)) = \text{Pr}_0(A_k|\mathbf{pa}(A_k))$, both of which have already been computed. Therefore we can calculate the CPT at $A_j$ as follows:

$$\frac{\text{Pr}_0(A_j|\mathbf{pa}(A_j))}{\text{Pr}_0(\overline{A_j}|\mathbf{pa}(A_j))}$$

$$= \frac{\text{Pr}_0(A_j|A_l \cup A_k \cup \mathbf{pa}(A_j))}{\text{Pr}_0(\overline{A_j}|A_l \cup A_k \cup \mathbf{pa}(A_j))} \cdot \frac{\text{Pr}_0(A_l|\tilde{\mathbf{pa}}(A_l))}{\text{Pr}_0(A_l|\mathbf{pa}(A_l))}$$

$$\cdot \frac{\text{Pr}_0(A_k|\tilde{\mathbf{pa}}(A_k))}{\text{Pr}_0(A_k|\mathbf{pa}(A_k))}, \qquad (13)$$

where $\tilde{\mathbf{pa}}(A_k)$ and $\tilde{\mathbf{pa}}(A_l)$ contain $\overline{A_j}$, and $\mathbf{pa}(A_k)$ and $\mathbf{pa}(A_l)$ contain $A_j$. Once we compute the likelihood ratio on the LHS of (13), the desired probabilities are uniquely determined, since $\text{Pr}_0(A_j|\mathbf{pa}(A_j)) + \text{Pr}_0(\overline{A_j}|\mathbf{pa}(A_j)) = 1$.

A consensus BN consistent with the LinOP would in general be fully connected, and thus not an object of particular value. However, if all agents' beliefs are given as BNs, we can retain their separation and still compute LinOP queries more efficiently. We exploit the fact that the LinOP obeys the marginalization property, and thus that the LinOP of any compound, marginal event can be computed as a LinOP over only that event. For example,

$$\text{Pr}_0(A_2\overline{A_5}A_9) = \sum_{i=1}^n \alpha_i \text{Pr}_i(A_2\overline{A_5}A_9),$$

where the terms on the RHS are calculated using a standard algorithm for Bayesian inference. Any conditional probability can be computed as the division of two compound, marginal probabilities.

Finally, we characterize the computational complexity of LinOP when all input models are BNs. Clearly, computing an arbitrary query $\text{Pr}_0(E|F)$ is NP-hard. Proposition 4 establishes that, even when all topologies agree, and even when only computing the LinOP of a CPT entry, the problem remains intractable.

**Proposition 4** *Let all input BNs have identical topologies. Then computing $\text{Pr}_0(A_j|\mathbf{pa}(A_j))$ consistent with LinOP is NP-hard.*

**Proof.** (sketch) Suppose that $n = 2$. Let $Pr_1$ be an arbitrary BN and let $Pr_2$ have an identical topology, but encode a uniform distribution—that is, $\text{Pr}_2(\omega) = 1/2^m$. We have shown that, if $\text{Pr}_0(A_m|\mathbf{pa}(A_m))$ were computable in polynomial time, then $\text{Pr}_1(A_m)$ could be inferred in polynomial time. Computing the later query is NP-hard [Cooper, 1990], and so the former must be as well. □

## 6 Conclusions

Graphical representations of a single agent's subjective belief form the core of many successful applications of uncertain reasoning. We examine the problem of combining several graphical models, to form a consensus model. Two intuitively reasonable assumptions in this context, made *a priori* by other authors, are (1) if all agents agree on a single topology, then that structure should be maintained, and (2) probability aggregation can be isolated within each conditional probability table (CPT). We demonstrate that each of these properties leads to an impossibility theorem when combined with other reasonable, oft-invoked assumptions. We prove that the logarithmic opinion pool (LogOP) maintains all agreed-upon Markov independencies, and describe procedures for constructing consensus Markov networks and consensus Bayesian networks that are consistent with the LogOP. We provide an algorithm for computing the CPTs of a LogOP-consistent consensus BN that takes advantage of available structure.

We consider the main contribution of this work to be an extension of known results on aggregating joint distributions to the case of combining graphical models. The results entail serious pitfalls for a modeler wishing to take into account the divergent opinions of multiple sources. Coherent combination of multiple models requires careful interpretation of the models to be combined, and deliberate consideration of the desired properties of the result.

**Acknowledgments**

Thanks to Eric Horvitz and the anonymous referees for insightful comments and pointers to related research. Part of this investigation was conducted while the first author was at Microsoft Research. Work at the University of Michigan was supported by AFOSR Grant F49620-97-0175.